\newcommand{\CLASSINPUTtoptextmargin}{0.75in}
\newcommand{\CLASSINPUTbottomtextmargin}{0.95in}
\newcommand{\CLASSINPUTbaselinestretch}{0.935}
\documentclass[conference]{IEEEtran}
\IEEEoverridecommandlockouts
\usepackage[style=ieee,doi=false,url=false,isbn=false]{biblatex}
\AtBeginBibliography{\footnotesize}
\usepackage{amsmath,amssymb,amsfonts}
\usepackage{algorithm}
\usepackage{algpseudocode}
\usepackage{graphicx}
\usepackage{textcomp}
\usepackage{xcolor}
\usepackage{acronym}
\usepackage{url}
\usepackage{microtype}
\usepackage{siunitx}
\usepackage{multirow}
\def\BibTeX{{\rm B\kern-.05em{\sc i\kern-.025em b}\kern-.08em
    T\kern-.1667em\lower.7ex\hbox{E}\kern-.125emX}}

\def\ve#1{\ensuremath{{\mathchoice{\mbox{\boldmath$\displaystyle #1$}}	%
			{\mbox{\boldmath$\textstyle #1$}}			%
			{\mbox{\boldmath$\scriptstyle #1$}}			%
			{\mbox{\boldmath$\scriptscriptstyle #1$}}}}}	%
\def\ma#1{\ensuremath{{\mathchoice{\mbox{\boldmath$\displaystyle #1$}}		%
			{\mbox{\boldmath$\textstyle #1$}}				%
			{\mbox{\boldmath$\scriptstyle #1$}}			%
			{\mbox{\boldmath$\scriptscriptstyle #1$}}}}}
\newcommand{\argmin}{\mathop{\mathrm{argmin}}}	

\newcommand{\norm}[1]{\ensuremath{ \Vert #1 \Vert}}
\newcommand{\RR}{\ensuremath{\mathbb{R}}}

\newcommand{\prox}[2]{\ensuremath{\mathrm{prox}_{#1}(#2)}}

\newlength{\bracewidth}
\newcommand{\myunderbrace}[2]{\settowidth{\bracewidth}{$#1$}#1\hspace*{-1\bracewidth}\smash{\underbrace{\makebox{\phantom{$#1$}}}_{#2}}}
\newcommand{\myoverbrace}[2]{\settowidth{\bracewidth}{$#1$}#1\hspace*{-1\bracewidth}\smash{\overbrace{\makebox{\phantom{$#1$}}}^{#2}}}

\newcommand{\gcal}{\ensuremath{\mathcal{G}}}
\newcommand{\Wlayer}[1]{\ensuremath{\ma{W}^{(#1)}}}

\newcommand{\Wlayertilde}[1]{\ensuremath{\Tilde{\ma{W}}^{(#1)}}}
\newcommand{\blayer}[1]{\ensuremath{\ve{b}^{(#1)}}}
\newcommand{\xlayer}[1]{\ensuremath{\ve{x}^{(#1)}}}
\newcommand{\Clayer}[1]{\ensuremath{\mathcal{C}^{(#1)}}}
\newcommand{\Ilayer}[1]{\ensuremath{\mathcal{I}^{(#1)}}}
\newcommand{\glayer}[1]{\ensuremath{\ve{g}^{(#1)}}}

\begin{document}
\acrodef{lcc}[LCC]{linear computation coding}
\acrodef{fs}[FS]{fully sequential}
\acrodef{fp}[FP]{fully parallel}
\acrodef{ua}[MA]{mixed algorithm}
\acrodef{dag}[DAG]{directed acyclic graph}
\acrodef{scm}[SCM]{single constant multiplication}
\acrodef{mcm}[MCM]{multiple constant multiplication}
\acrodef{cmvm}[CMVM]{constant matrix vector multiplication}
\acrodef{dmp}[DMP]{discrete matching pursuit}
\acrodef{rs}[RS]{reduced state}
\acrodef{lz}[LZ]{Lempel-Ziv}
\acrodef{sqnr}[SQNR]{signal to quantization noise ratio}
\acrodef{csd}[CSD]{canonically signed digit}
\acrodef{fpga}[FPGA]{field programmable gate array}
\acrodef{asic}[ASIC]{application-specific integrated circuit}
\acrodef{lut}[LUT]{lookup table}
\acrodef{nn}[NN]{neural network}
\acrodef{ls}[LS]{least squares}
\acrodef{cse}[CSE]{common subexpression elimination}
\acrodef{mlp}[MLP]{multilayer perceptron}
\acrodef{fk}[FK]{full kernel}
\acrodef{pk}[PK]{partial kernel}
\acrodef{ck}[CK]{custom kernel}
\acrodef{resnet}[ResNet]{Residual Net}

\title{Coding for Computation: Efficient Compression of Neural Networks for Reconfigurable Hardware\\
\thanks{This work was supported by Deutsche Forschungsgemeinschaft (DFG) under the project Computation Coding (MU-3735-/8-1). The authors gratefully acknowledge the scientific support and HPC resources provided by the Erlangen National High Performance Computing Center (NHR@FAU) of the Friedrich-Alexander-Universität Erlangen-Nürnberg (FAU). The hardware is funded by the DFG.}
}

\author{\IEEEauthorblockN{Hans Rosenberger\IEEEauthorrefmark{1}, Rodrigo Fischer\IEEEauthorrefmark{2}, Johanna S. Fröhlich\IEEEauthorrefmark{1}, Ali Bereyhi\IEEEauthorrefmark{3}, Ralf R. M\"uller\IEEEauthorrefmark{1}}
\IEEEauthorblockA{\IEEEauthorrefmark{1}Institute for Digital Communications, Friedrich-Alexander-Universität (FAU), Erlangen, Germany \\
%\{hans.rosenberger, ralf.r.mueller\}@fau.de \\
\IEEEauthorrefmark{2}Communications Engineering Lab, Karlsruhe Institute of Technology (KIT), Karlsruhe, Germany \\
\IEEEauthorrefmark{3}Electrical and Computer Engineering Department, University of Toronto, Canada 
}
}

\maketitle

\begin{abstract}
As state of the art \acp{nn} continue to grow in size, their resource-efficient implementation becomes ever more important.
In this paper, we introduce a compression scheme that reduces the number of computations required for \ac{nn} inference on reconfigurable hardware such as FPGAs.
This is achieved by combining pruning via regularized training, weight sharing and \ac{lcc}.
Contrary to common \ac{nn} compression techniques, where the objective is to reduce the memory used for storing the weights of the \acp{nn}, our approach is optimized to reduce the number of additions required for inference in a hardware-friendly manner. 
%The proposed scheme is better than alternative methods for simple multilayer perceptrons, as well as for large scale deep \acp{nn} such as ResNets.
The proposed scheme achieves competitive performance for simple multilayer perceptrons, as well as for large scale deep \acp{nn} such as ResNet-34.
\end{abstract}

%\begin{IEEEkeywords}
%component, formatting, style, styling, insert
%\end{IEEEkeywords}

\acresetall
\vspace{-2mm}
\section{Introduction}
Deep \acp{nn} have shown remarkable success across a wide range of applications, including computer vision~\cite{He2015,Tan2019}, machine translation~\cite{Vaswani2017}, and speech recognition~\cite{Graves2013}.
Their computational complexity and memory requirements, however, pose significant challenges for their deployment.
Most research on \ac{nn} compression focuses on reducing the storage requirements of a given model. This is particularly beneficial when deploying networks on resource-constrained devices or transmitting them to edge devices~\cite{Murshed2021}.

This paper adopts a different approach for \ac{nn} compression.
Rather than minimizing the number of bits required to represent the model parameters, our objective is to reduce the number of additions necessary for computing matrix-vector products, which are an essential operation for inference in \acp{nn}.
This is particularly advantageous when deploying \acp{nn} on reconfigurable hardware, such as \acp{fpga}, which are already deployed as \ac{nn} accelerators in data-centers~\cite{Yu2019}; see for instance Microsoft's Project Brainwave~\cite{Chung2018}.

%our goal is not to minimize the number of bits to represent the \ac{nn}, but to minimize the number of additions required in computing the matrix vector products, that are a fundamental part of common \ac{nn} layers. 
%This is particularily useful if a \ac{nn} is deployed on reconfigurable hardware, such as \acp{fpga}.
%So far, \acp{fpga} have been successfully used as \ac{nn} accelerators in data centers (e.g. Microsoft Project Brainwave)~\cite{Chung2018,Yu2019}.

The paper is structured as follows: Section~\ref{sec::preliminaries} presents the fundamental concepts and reviews related work. Section~\ref{sec::nncompression} outlines our proposed compression scheme, while Section~\ref{sec::numericalresults} evaluates its performance through numerical simulations.

%\subsection{Contributions}

\subsection{Notation}
Vectors, matrices and sets are denoted by lowercase boldface letters, uppercase boldface letters and uppercase caligraphic letters, i.e. $\ve{x}$, $\ma{X}$ and $\mathcal{X}$, respectively.
The $i$-th row of a matrix $\ve{X}$ is shown as $[\ve{X}]_i$.
The cardinality of a set is denoted as $|\mathcal{X}|$ and $[N]$ represents the set $\{ 1,\dots,N \}$.
The $p$-norm and the matrix transpose are shown by $\norm{\cdot}_p$ and $(\cdot)^\mathrm{T}$, respectively.
%We use the abbreviation $(\cdot)_+$ to denote $\max(\cdot,0)$.
%The augmented identity matrix with dimension $N \times K$ is denoted by $\ma{I}_{N \times K}$.
The gradient of a function $f$ with respect to a matrix $\ma{X}$ is denoted as $\nabla_{\ma{X}} f$.
\vspace{-1mm}
\section{Preliminaries}\label{sec::preliminaries}
Consider a general feed-forward \ac{nn}, where each layer can be expressed in the form
\begin{align}
    \ve{y} = f \left( \ma{W}\ve{x} + \ve{b} \right).
\end{align}
Here, $\ve{x} \in \RR^K$ is an input feature, $\ma{W} \in \RR^{N \times K}$ is a weight matrix, $\ve{b} \in \RR^N$ is a bias vector, $\ve{y} \in \RR^N$ the layer output and $f(\cdot):\RR \rightarrow \RR$ is a nonlinear activation function acting elementwise on the components of a vector. 
Our objective is to reduce the number of operations required to compute the matrix-vector product $\ma{W}\ve{x}$.
Given a finite-precision/quantized representation of $\ma{W}$, evaluating $\ma{W}\ve{x}$ consists of additions/subtractions, along with multiplications by signed powers of two.
The latter can be considered computationally cheap, as multiplications by powers of two correspond to bitshifts in hardware.

To illustrate the point mentioned above, let us consider a simple example:
\vspace{2mm}
\begin{align}\label{example:matprod}
    \underbrace{\begin{pmatrix}
        2 & 0.375 \\
        3.75 & 1 
    \end{pmatrix}}_{\ma{W}}
    \underbrace{\begin{pmatrix}
        x_1 \\ x_2
    \end{pmatrix}}_{\ve{x}} 
    = 
    \begin{pmatrix}
        \myoverbrace{2^{1} x_1 + 2^{-1} x_2}{m(x_1,x_2)} - 2^{-3} x_2 \\
        - 2^{-2} x_1 + \myunderbrace{2^2 x_1 + 2^{0} x_2}{2m(x_1,x_2)} \\
    \end{pmatrix}
\end{align}
Computing the product of this example requires two additions, two subtractions and six bitshifts.
To reduce the number of additions/subtractions in this product, we have two options:
\begin{itemize}
    \item \textit{Remove unnecessary or less important entries in $\ma{W}$}: 
    Various approaches have been explored to reduce the number of parameters in \acp{nn}, including knowledge distillation~\cite{Hinton2015}, weight-sharing~\cite{Nowlan1992,Ullrich2017}, low-rank factorizations~\cite{Denton2014,Jaderberg2014} and network pruning~\cite{blalock:surveypruning:20,Karnin_1990,LeCun1989}.
    For the purpose of minimizing parameters of a given network structure, pruning is the most straightforward choice.
    Specifically, one can employ pruning through regularization (c.f. Section~\ref{sec::regtraining}).
    Subsequent to pruning, weight sharing can be utilized via a retraining procedure to tie similar parameters to the same value (c.f. Section~\ref{sec::weightsharing}).
    \item \textit{Remove redundant computations in $\ma{W}\ve{x}$}: 
    In a matrix-vector product, certain subexpressions may appear redundantly. 
    For instance, in~\eqref{example:matprod}, the term $m(x_1,x_2)$ appears twice, differing only by a factor of two (i.e. a bitshift). 
    By computing $m(x_1,x_2)$ first and reusing it in the remaining computation, one addition can be eliminated. 
    Several techniques exist for identifying such redundancies, including \ac{cse}~\cite{Kamal2017}, \ac{dag} based methods~\cite{Voronenko_2007}, digit recoding approaches~\cite{Oudjida_2016} as well as hybrid methods~\cite{Aksoy_2015}.
    However, many of these approaches face significant challenges: either the optimization problems required to detect redundancies are computationally prohibitive for the large matrix sizes typically encountered in \acp{nn}, or they focus on restricted subproblems, such as \ac{mcm}.

    To address these limitations, we employ an information-theoretic framework for \ac{cmvm}, known as \ac{lcc}, which remains efficient even for large matrices (cf.~ Section~\ref{sec::lcc}).
    
    %Within a product of a constant matrix with an arbitrary vector there exist subterms that are redundant. A toy example can be found in~\eqref{example:matprod} where the term $m(x_1,x_2)$ can be found in two separate instances only differing by a factor of two, i.e. a bitshift. Computing $m(x_1,x_2)$ first and then the remainder of the matrix vector product thus saves one addition. Methods explicitly searching for these common terms are called \ac{cse} methods~\cite{Kamal2017}. Other methods include \ac{dag} based schemes~\cite{Voronenko_2007}, digit recoding based approaches~\cite{Oudjida_2016} or combinations thereof~\cite{Aksoy_2015}.
\end{itemize}
By choosing complementary methods for both strategies, we not only reduce the number of additions for each approach but also achieve an additional \textit{combining gain}, as will be illustrated in the forthcoming sections.
%By selecting complementary methods for both strategies, we aim to achieve not only a reduction in the number of additions for each technique but also, in many cases as we will observe in the sequel, an additional \textit{combining gain}.

\section{Neural Network Compression for Computation}\label{sec::nncompression}
In this section, we present our compression scheme by first examining its individual components, followed by an outline of the complete procedure.
\subsection{Linear Computation Coding}\label{sec::lcc}
A particularly efficient approach to remove redundant computations in \acp{cmvm} is \ac{lcc}.
For processing by \ac{lcc} tall matrices are desirable, ideally with an exponential aspect ratio. 
If matrices are wide or square, we vertically slice them into multiple tall submatrices~\cite{Lehnert_2023}, i.e.
\begin{align}
    \ma{W} = [\ma{W}_1|\ma{W}_2|\cdots|\ma{W}_E].
\end{align}
The key idea is then to express each of the slices as a product of matrices, i.e.
\vspace{-2mm}
\begin{align}
    \ma{W}_e \approx \ma{F}_{e,P} \cdots \ma{F}_{e,1} \ma{F}_{e,0}
\end{align}
where each of the matrix factors $\ma{F}_{e,p}$ has a well-defined structure and each row only contains powers of two or zeros~\cite{M_ller_2022}.

There exist multiple algorithms for obtaining these sparse matrix factors.
Due to space constraints, we refer the reader to~\cite{Rosenberger2024} for a detailed discussion of \ac{lcc} algorithms.
Here, we provide only a brief high-level overview of two key algorithms used in our subsequent analysis:
\begin{itemize}
    \item \Ac{fp} algorithm: 
    Each row in every matrix factor $\ma{F}_{e,p}$ contains at most $S$ nonzero elements, all of which are signed powers of two.
    The primary advantage of this approach is that computing the matrix-vector product requires no more than $S-1$ additions per row.
    Additionally, the computations for each of the $N$ rows are independent.
    %The primary advantage of this approach is that computing each matrix factor at most $N (S-1)$ additions are required, and each of the $S-1$ computations of the $N$ rows of $\ma{F}_{e,p}$ is fully independent.
    This independence makes the \ac{fp} algorithm highly suitable for parallel execution on reconfigurable hardware such as \acp{fpga}.
    However, its performance can degrade when matrices are small or not well-behaved\footnote{We consider matrices as not well-behaved, if their row vectors only span a subspace of the full space $\RR^K$.}, making it less efficient in such cases.
    \item \Ac{fs} algorithm: 
    The major disadvantage of the \ac{fp} algorithm is alleviated by the \ac{fs} algorithm, which does not require computations at each step to be independent. 
    The graph of the computations between input and output of the matrix-vector product is not structured anymore. 
    Hence, while offering a better performance for arbitrary types of matrices, the \ac{fs} algorithm may not be as suited for parallelization as the \ac{fp} algorithm.
\end{itemize}

%he matrix factors are obtained recursively via
%\begin{align}\label{eq::sparserecovery}
%    [\ma{F}_{e,p}^\mathrm{T}]_i = \underset{\ve{\omega}\in\mathcal{W}_S}{\argmin} \norm{[\ma{W}_e^\mathrm{T}]_i - \ve{\omega}\ma{F}_{e,p-1}\dots\ma{F}_{e,0}} \quad \forall e \in [E],
%\end{align}
%with
%\begin{align}
%    \mathcal{W}_S = &\biggl\{ \ve{\omega} = \sum_{s=1}^S i_s \ve{1}_{j_s, \, L} : i_s \in \mathcal{M} \subseteq \{ 0, \pm 2^\ZZ \}, j_s \in \shortset{L} \biggr\}.
%\end{align}
%contains only vectors with at most $S$ elements that are powers of two.
%Consequently, each row in $\ma{F}_{e,p}$ requires than $S-1$ additions or subtractions

The following properties of \ac{lcc} influence its effectiveness:
\begin{itemize}
    \item \Ac{lcc} works best for matrices with an exponential aspect ratio~\cite{M_ller_2022}.
    %The taller the matrix, the greater the gain obtained by \ac{lcc}~\cite{M_ller_2022}. 
    Consequently, reducing the number of input neurons can, in many cases, help skew the aspect ratio of a weight matrix in favor of \ac{lcc}, provided that the output dimension remains fixed.
    \item \Ac{lcc} is most effective for dense matrices. 
    Unstructured sparsity, i.e. the arbitrary removal of individual weights, can degrade performance significantly. 
    To avoid a loss in performance, sparsity should be introduced in a structured manner, by removing entire rows or columns rather than individual entries. 
    %\textbf{TODO: If space allows in the end we can add simulation results that back up this claim.}
    %\item \textbf{Consider removing the rest}
    %\item Each matrix has a well defined structure, containing only a fixed number of powers of two. Consequently, each matrix factor has a fixed number of additions~\cite{M_ller_2022}.
    %\item The \ac{fp} algorithm is ideally suited for \textit{pipelining} on reconfigurable hardware such as \acp{fpga}~\cite{Lehnert_2023}.
    %\item Obtaining \ac{lcc} decomposition via the \ac{fp} or \ac{fs} algorithm is computationally tractable, even for large weight matrices often found in \acp{nn}.
\end{itemize}

\subsection{Regularization via Group Lasso}\label{sec::regtraining}
For ease of notation, we write the parameters of the \ac{nn}, namely the set of weight matrices and bias vectors, as $\gcal = \{ \Wlayer{1}, \blayer{1}, \dots, \Wlayer{L}, \blayer{L} \}$, where the superscript $(l)$ denotes the layer index $l \in [L]$.
The training of a \ac{nn} can then be expressed as an optimization problem
\begin{align}\label{eq::trainingobj}
    \underset{\gcal}{\min} \quad L(\gcal) + R(\gcal),
\end{align}
where $L(\cdot)$ represents the loss function, such as the cross-entropy or Kullback-Leibler divergence~\cite{goodfellow2016deep} and $R(\cdot)$ is a regularization term to facilitate the removal of unnecessary or less significant weights.

For the purpose of pruning, the regularizer can be set to the $\ell_1$-norm~\cite{Collins2014}.
The $\ell_1$-regularization, often called lasso, results in sparse weight matrices.
Though reducing the number of non-zero entries, the resulting sparsity of the matrices can severely degrade the performance of \ac{lcc} algorithms.
%However, weight matrices are sparse after training rendering the performance of \ac{lcc} algorithms severely degraded.
To incorporate this regularization into \ac{lcc}, we need to remove unimportant weights while retaining dense weight matrices.
One approach is to use a group lasso penalty, where entire \textit{groups} of weights, rather than individual weights, are penalized~\cite{Scardapane2017}.
The regularization objective then reads
\begingroup
\setlength{\belowdisplayskip}{1pt}
\begin{align}\label{eq::grouplasso}
    R(\gcal) = \sum_{l \in [L]} \underbrace{\lambda_{l,1} \sum_i \norm{[\Wlayertilde{l}]_i}_2}_{:=r_l^\mathrm{GrL}(\Wlayertilde{l})}, %+ \lambda_{2,l} \norm{\Wlayer{l}}_\mathrm{F}^2
\end{align}
\endgroup
where $\Wlayertilde{l}$ is a reshaped version of $\Wlayer{l}$, such that each row in $\Wlayertilde{l}$ corresponds to a group that we either want to retain or prune and $\lambda_{l,1}$ is the regularization hyperparameter.
Suitable groups are in general a design choice: common choices are individual filters, neurons, etc.~\cite{Wen2016}.
For dense layers, we want to remove input neurons/columns of $\Wlayer{l}$.
Hence, we set $\Wlayertilde{l} = (\Wlayer{l})^\mathrm{T}$.
Suitable choices for groups in convolutional layers are discussed in Section~\ref{sec::convlayers}.

As the regularization objective in~\eqref{eq::grouplasso} is convex but nondifferentiable, we resort to a proximal gradient algorithm to minimize~\eqref{eq::trainingobj}~\cite{Hastie2015Lasso}, i.e.
%\vspace{-1mm}
\begin{align}
    \Wlayer{l} \gets \prox{\eta r_l^\mathrm{GrL}}{\Wlayer{l}-\eta \nabla_{\Wlayer{l}} L(\gcal)},
\end{align}
where $\eta$ is the learning rate.
We further recall the definition of the proximal operator of a scaled convex function $\lambda g$ as $\prox{\lambda g}{v} = \argmin_x g(x) + 1/(2\lambda) \norm{x-v}_2^2$~\cite{boyd2014proximal}.
Given some matrix $\ma{A} \in \RR^{I \times M}$, the proximal operator of $r_l^\mathrm{GrL}(\ma{A})$ is equivalent to performing block soft thresholding on the individual rows $[\ma{A}]_i$~\cite{Hastie2015Lasso}, i.e.
%\vspace{-1mm}
\begin{align}\label{eq::blocksoftthres}
    \prox{\eta r_l^\mathrm{GrL}}{\ma{A}} := \left[ \max \left( 1-\frac{\eta\lambda_{l,1}}{\norm{[\ma{A}]_i}_2}, 0 \right) \! [\ma{A}]_i \right]_{1 \leq i \leq I}.
\end{align}

\subsection{Weight Sharing}\label{sec::weightsharing}
After training, some columns of the weight matrices can be highly correlated.
To further reduce computations, it is desirable that clusters of similar columns share the same values, provided this has minimal impact on the prediction accuracy of the given \ac{nn}.
To determine suitable clusters of similar columns and replace them by their centroids via a retraining procedure, we adopt the method proposed in~\cite{zhang2018learning}, which is outlined in the sequel.

For each weight matrix $\Wlayer{l}$, we determine clusters of highly correlated columns by applying the affinity propagation algorithm~\cite{Frey2007} built into the \textit{scikit-learn} package~\cite{scikit-learn_2011} in Python.
Contrary to other clustering algorithm such as k-means, affinity propagation does not require a prior specification of the number of clusters.

During retraining, let $\Clayer{l}_i$ denote the $i$-th cluster\footnote{$\Clayer{l}_i$ contains all of the columns of $\Wlayer{l}$ associated with the cluster.} of layer $l$; the centroid $\glayer{l}_i$ associated with this cluster is updated via~\cite{zhang2018learning}
\begin{align}\label{eq::centroids}
    \frac{\partial L}{\partial \glayer{l}_i} = \frac{1}{\left| \Clayer{l}_i \right|} \sum_{\ma{w} \in \Clayer{l}_i} \frac{\partial L}{\partial \ma{w}}.
\end{align}

After the retraining procedure is concluded, the columns associated with each cluster $\Clayer{l}_i$ are replaced by their corresponding centroids $\glayer{l}_i$.
Consequently, we can replace the weight matrix $\Wlayer{l}$ after retraining, by a matrix containing only all of the unique cluster centroids $\glayer{l}_i$, i.e.
\begin{align}
    \Wlayer{l} \xlayer{l} &= 
    %\sum_k [\Wlayer{l}]_k x_k^{(l)} = 
    \sum_i \glayer{l}_i \sum_{j \in \Ilayer{l}_i} x_j^{(l)},
\end{align}
where $\Ilayer{l}_i$ contains the column indices of the columns in $\Clayer{l}_i$.
Note that summing over the elements of $\xlayer{l}$ corresponding to each centroid only requires scalar additions.

\subsection{Convolutional Layers}\label{sec::convlayers}
In addition to fully-connected layers, convolutional layers are another common form of layers in feed-forward architectures.
In convolutional layers, multiple kernels (filters) are convolved with the input to extract spatial features.
We consider convolutional layers where each of the $K$ input feature maps of size $Z \times Z$ is convolved with $N$ individual kernels of size\footnote{We assume that kernels are square as it is often the case. However, our scheme works for rectangular kernels as well.} $O \times O$.
%With a slight abuse of notation, the weight matrix of a convolutional layer can be expressed as $\Wlayer{l} \in \RR^{O \times O \times N \times K}$, where $O \times O$ denotes the individual dimension of a kernel\footnote{We assume for simplicity that kernels are square as it is often the case, however the following methods work as well for any type of rectangular kernel.}, $N$ represents the number of output features and $K$ the number of input features.

\begin{figure}[!t]
    \centering
    \includegraphics[width=\linewidth]{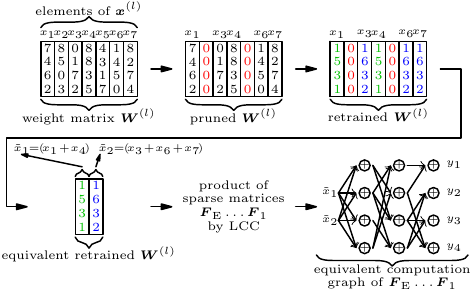}
    \caption{Schematic overview of the compression scheme detailed in Section~\ref{sec::nncompression}.}
    \label{fig:scheme_comp}
\end{figure}

From a computational perspective, a convolutional layer can be equivalently reformulated as collection of matrix-vector products.
Two representations are outlined in~\cite{Mueller_2023}:
\begin{itemize}
    \item \textit{\Ac{fk} method:}
    The convolutional layer can be interpreted as a collection of $K$ matrices, where each matrix performs $N$ parallel convolutions corresponding to a single input feature map.
    Let $\ve{x}_{k}$ be the local receptive field of the $k$-th input feature map, then the $N$ parallel convolutions can be computed via $\ve{y}_k = \Wlayer{l}_k \ve{x}_k$, where each element in $\ve{y}_k$ corresponds to the result of an individual convolution and the rows of  $\Wlayer{l}_k \in \RR^{N \times O^2}$ contain the individual flattened convolutional filters.
    
    %Hence, the matrix corresponding to the $k$-th input feature map is of the form $\Wlayer{l}_k \in \RR^{N \times O^2}$.
    
    %Given an input feature of size $Z \times Z$, the matrix vector product $\ve{y}_j = \Wlayertilde{l}_k \ve{x}_j$ has to be evaluated $(Z-O+1)^2$ times, where the $\ve{x}_j$ corresponds to the local receptive field\footnote{Current subarea of the input feature that the kernel is convolved with.} of the input feature currently being convolved with the $k$-th matrix.

    Concerning the regularization objective, each kernel is treated as a group, i.e., the rows of $\Wlayer{l}_k$ are considered groups:
    \vspace{-2mm}
    \begin{align}\label{eq::convobj}
    \Wlayertilde{l} = [(\Wlayer{l}_1)^\mathrm{T} \cdots (\Wlayer{l}_K)^\mathrm{T} ]^\mathrm{T}.
    \end{align}
%the $N$ parallel convolutions for each of the $K$ input features as a matrix vector product $\ve{y} = \Wlayertilde{l} \ve{x}$, where $\Wlayertilde{l}_k \in \RR^{N \times O^2}$ is the equivalent reshaped matrix corresponding to the $k$-th input features. The input vector $\ve{x}$ is the current flattened (sub-)area of the input feature being convolved with the $N$ kernels in parallel.
    \item \textit{\Ac{pk} method:} 
    The spatial dependencies of convolutional filters enable alternative representations that are more suitable for \ac{lcc}, i.e. leading to taller matrices.

    We restructure the weight matrix of the $k$-th input feature to: $\Wlayer{l}_k \in \RR^{NO \times O}$, where each row corresponds to a single column or row\footnote{For the numerical simulations we will use columns of filters as groups, however the approach works equally well for rows.} of a kernel.
    Regarding the regularization objective, we can apply~\eqref{eq::convobj} analogously to the restructured matrices $\Wlayer{l}_k$ of the \ac{pk} method.
 
    After computing the matrix-vector product for all possible kernel positions on each input feature, the resulting $O$ partial outputs for each convolution must be summed to obtain the final output.
\end{itemize}
Various other strategies can be used to restructure convolutions into matrix-vector products. 
However, exploring these methods is beyond the scope of this paper.

The overall training procedure and decomposition by \ac{lcc} is summarized in Algorithm~\ref{alg:overallalg} and a schematic overview of our proposed compression scheme is depicted in Fig.~\ref{fig:scheme_comp}.

%\noindent\rule{0pt}{1cm} 
\begin{algorithm}[b]
    \caption{\ac{nn} compression procedure for \ac{lcc}}
    \begin{algorithmic}
        \Require Model parameters $\gcal$ initialized for training, hyperparameters $\{ \lambda_{1,1}, \dots, \lambda_{L,1} \}$, learning rate $\eta$
        \For{epoch $=1,\dots$} \Comment{Regularized training procedure}
            \State Train model parameters in $\gcal$ via backpropagation
            \For{$l=1,\dots,L$}
                \State Apply proximal operator~\eqref{eq::blocksoftthres} to weight matrix $\Wlayertilde{l}$
            \EndFor
        \EndFor
        \State Apply retraining procedure to obtain clusters and their corresponding centroids via~\eqref{eq::centroids}
        \For{$l=1,\dots,L$} \Comment{\Ac{lcc} decomposition}
            \If{$\Wlayer{l}$ is dense}
                \State Apply \ac{lcc} algorithm to $\Wlayer{l}$
            \ElsIf{$\Wlayer{l}$ is convolutional}
                \State Apply \ac{lcc} algorithm to each $\Wlayer{l}_k$
            \EndIf
        \EndFor
        \State \Return $\gcal$, \ac{lcc} decompositions
    \end{algorithmic}
    \label{alg:overallalg}
\end{algorithm}

\section{Numerical Evaluation}\label{sec::numericalresults}
We now validate the effectiveness of our proposed scheme by using it to compress a simple \ac{mlp} and ResNet-34.

To assess the performance of our compression scheme, we focus on the tradeoff between prediction accuracy and compression ratio. 
We define the compression ratio as the number of additions of the uncompressed model relative to the number of additions of the compressed model. 
To establish a baseline, we train the uncompressed model without regularization and determine the number of additions using the \ac{csd} representation~\cite{Booth_1951}.

For simplicity, our analysis considers only the additions required for matrix-vector products within each network, excluding other computational costs associated with \ac{nn} inference, such as activation functions. 
Prediction accuracy is quantified using the top-1 accuracy.

%To quantify the compression yielded by the regularized training and weight-sharing procedure, we define the compression ratio as number of unique columns/number of total parameters \footnote{In several publications such as~\cite{zhang2018learning} the compression ratio is defined as number of unique parameters/number of total parameters. However we found that after regularized training it is in many instances likely that multiple values within the matrix posses the same value, which artificially inflates compression. 
%Further, to compress matrices by column weight sharing for subsequent processing by \ac{lcc} the number of unique columns is the relevant metric.
%}.
%For all models decomposed by \ac{lcc} we define as the compression ratio the number of additions for the compressed model/number of additions for the uncompressed model\footnote{For the number of additions we use the \ac{csd} representation~\cite{Booth_1951} as a baseline.}. 

%The code used to reproduce the numerical results can be found in our \textit{GitHub} repository: \url{https://github.com/hansrosenberger/computationcoding}.
%\textbf{TODO: Check if it makes sense to publish the code in our repo.}

\subsection{\ac{mlp} trained on MNIST}
As the first experiment, we consider an \ac{mlp} with one hidden layer of width 300 trained for MNIST classification~\cite{lecun2010mnist}. 
%\textbf{TODO: We could have here also results for fashion MNIST, etc. However i think as space is very limited is something more for the extended version.}
%The dataset contains grayscale images of size $28\times28$ (784 pixels) that belong to either one of ten classes.
We train the \ac{mlp} for 200 epochs using stochastic gradient descent with an initial learning rate $\eta=0.001$ and a momentum of 0.9. 
We decrease the learning rate every ten epochs by a factor of 0.95.
We regularize the first layer and leave the second layer unregularized.

\begin{figure}[!t]
    \centering
    \includegraphics[width=\linewidth]{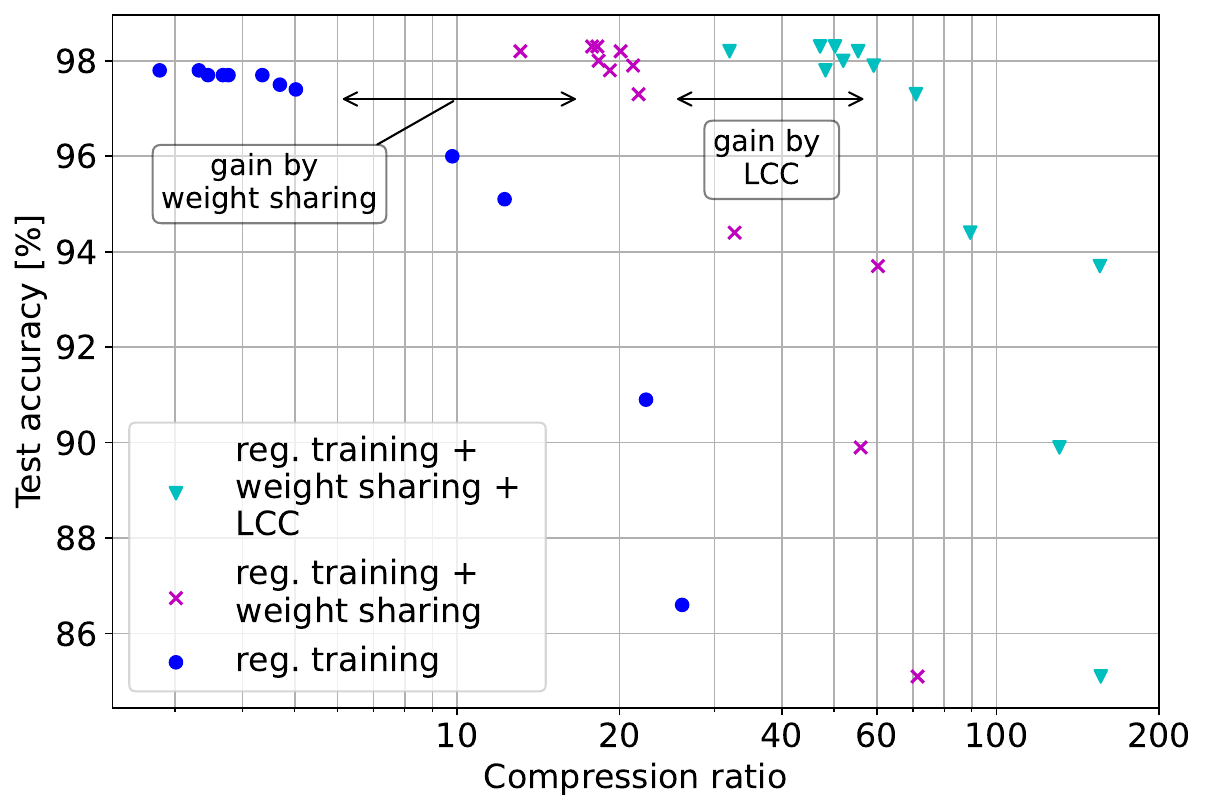}
    \caption{Compression-accuracy tradeoff for the first layer of the \ac{mlp} for various regularization parameters $\lambda_{1,1}$. Dots denote trained networks that were only compressed by regularized training/pruning, crosses denote compression by regularized training and subsequent weight sharing and triangles denote the full compression scheme (regularized training, weight sharing and decomposition by \ac{lcc}).}
    \label{fig:mnist_results}
\end{figure}

Fig.~\ref{fig:mnist_results} illustrates the compression-accuracy tradeoff for different regularization parameters $\lambda_{1,1}$ after training the \ac{mlp}.
It is evident that each component of our proposed compression scheme enhances compression while maintaining minimal impact on prediction accuracy.
The increase in compression when only considering the contribution of \ac{lcc} ranges from a factor of 2.4 to 3.1, primarily due to the transformation of the equivalent retrained weight matrices into tall matrices through pruning and weight sharing.
Specifically, the matrix dimensions are reduced from $300 \times 784$ (without pruning) to between $300 \times 14$ and $300 \times 45$, significantly improving \ac{lcc} efficiency.
If we apply \ac{lcc} directly to the unpruned weight matrix without weight sharing the compression ratio would only increase by a factor of two.
Thus, integrating \ac{lcc} with pruning via regularized training and weight sharing results in an additional performance gain of up to \SI{50}{\percent}.

%We can clearly observe, that each individual component of our proposed compression scheme increases compression, while keeping the impact on prediction accuracy minimal.
%The individual gain achieved by \ac{lcc} lies between $2.4-3.1$, this is mainly achieved as the equivalent retrained weight matrices are tall through the pruning and weight sharing procedure, i.e. of dimension $300 \times 14$ to $300 \times 45$.
%Without pruning the weight matrix of the first layer has dimension $300 \times 784$ and \ac{lcc} would only increase compression by a factor of 2.
%Hence, we can conclude that in the given case combining \ac{lcc} with pruning through regularized training and weight sharing leads to an additional performance gain of up to $\SI{50}{\percent}$.

\subsection{Experiments with ResNet-34}
To evaluate the performance of our approach for more realistic scenarios, we consider a residual network (ResNet)~\cite{He2015}, namely ResNet-34, with the improved bottleneck blocks as proposed in~\cite{He2016}. 
The TinyImageNet dataset~\cite{Le2015}, a subset of the original ImageNet dataset~\cite{Deng2009}, is used for training and testing. 
This dataset consists of $10^6$ images evenly distributed across 200 classes.
We train the model using the Adam optimizer with an initial learning rate of 0.01.

Table~\ref{tab:resnet_results} presents the compression-accuracy tradeoff for ResNet-34 after pruning via regularized training and applying \ac{lcc}. 
The results indicate that the \ac{fs} algorithm, for both the \ac{fk} and \ac{pk} methods, achieves at least a twofold reduction in the number of additions while preserving prediction accuracy. 
In contrast, the \ac{fp} algorithm yields only marginal gains. 
This can be attributed to the high compression ratio already achieved through regularized training, which results in relatively small equivalent weight matrices for \ac{lcc} decomposition. 
In such cases, the \ac{fs} algorithm proves to be a more effective choice compared to the \ac{fp} algorithm.

\begin{table}[!t]
    \centering
    \caption{Compression-accuracy results for ResNet-34 for regularized training and subsequent decomposition by different \ac{lcc} algorithms and kernel representations.
    The baseline accuracy for TinyImageNet on ResNet-34 without regularization is \SI{59.0}{\percent}.}
    \label{tab:resnet_results}
    \begin{tabular}{|c|c|c|}
        \hline
        Method & \ac{fk} method & \ac{pk} method \\\hline\hline
        \multirow{2}{*}{reg. training} & 22.8 & 21.4 \\
        & \SI{55.2}{\percent} & \SI{57.0}{\percent} \\\hline
        reg. training + & 25.2 & 22.7 \\
        \ac{lcc} (\ac{fp} algorithm) & \SI{55.2}{\percent} & \SI{57.0}{\percent} \\\hline
        reg. training + & 46.5 & 43.9 \\
        \ac{lcc} (\ac{fs} algorithm) & \SI{55.2}{\percent} & \SI{56.9}{\percent} \\\hline
    \end{tabular}
\end{table}

\section{Summary \& Conclusion}
This work proposed a novel \ac{nn} compression scheme that minimizes the number of additions in matrix-vector multiplications across the layers of a deep \ac{nn}.
Our approach delivers substantial computational savings, demonstrating effectiveness for inference in deep \acp{nn}.

%\bibliographystyle{ieeetr}
%\bibliography{IEEEabrv,jabref_collection}
\printbibliography

\end{document}